\documentclass[letterpaper, 10 pt, conference]{IEEEtran}  % Comment this line out if you need a4paper

\IEEEoverridecommandlockouts                            

\usepackage{graphics} % for pdf, bitmapped graphics files
\usepackage{epsfig} % for postscript graphics files
\usepackage{times} % assumes new font selection scheme installed
\usepackage{amsmath} % assumes amsmath package installed
\usepackage{amssymb}  % assumes amsmath package installed
\usepackage{color}
\usepackage{algorithm}%
\usepackage{algorithmicx}%
\usepackage{algpseudocode}%
\usepackage{threeparttable}
\usepackage{multirow}
\usepackage{graphicx}
\usepackage{subfigure}

\usepackage{cite}
\usepackage[dvipsnames]{xcolor}
\usepackage{adjustbox}
\usepackage{array}
\usepackage{makecell}
\usepackage[hidelinks]{hyperref}
\usepackage{booktabs}
\usepackage{url}

\usepackage{tikz}
\usetikzlibrary{shapes, arrows.meta, positioning, calc, fit, backgrounds, shapes.geometric}

\definecolor{netblue}{RGB}{230, 240, 255}
\definecolor{netborder}{RGB}{70, 130, 180}
\definecolor{lossred}{RGB}{255, 235, 235}
\definecolor{lossborder}{RGB}{200, 100, 100}
\definecolor{p1color}{RGB}{0, 114, 178}    % Blue (Phase 1: Data)
\definecolor{p2color}{RGB}{213, 94, 0}     % Orange (Phase 2: Ridge)
\definecolor{bothcolor}{RGB}{140, 40, 160} % Violet (Both: Physics)
\definecolor{paleorange}{rgb}{0.85, 0.45, 0.1} % darker orange-brown
\definecolor{skyblue}{rgb}{0.1, 0.45, 0.8}     % darker blue
\definecolor{palegreen}{rgb}{0.1, 0.6, 0.3}    % darker green

\begin{document}

\title{\LARGE \bf
}
\title{SOLIS: Physics-Informed Learning of Interpretable Neural Surrogates for Nonlinear Systems
\author{Murat Furkan Mansur and Tufan Kumbasar}% <-this % stops a space
\thanks{This work was supported by the project (125E231) of the Scientific and Technological Research Council of Türkiye (TUBITAK) }
\thanks{Murat Furkan Mansur and Tufan Kumbasar are with AI and Intelligent Systems Laboratory, Istanbul Technical University, 34469, Istanbul, Türkiye  {\tt\small mansur17@itu.edu.tr, kumbasart@itu.edu.tr}}%
}

\maketitle

\begin{abstract}
Nonlinear system identification must balance physical interpretability with model flexibility. Classical methods yield structured, control-relevant models but rely on rigid parametric forms that often miss complex nonlinearities, whereas Neural ODEs are expressive yet largely black-box. Physics-Informed Neural Networks (PINNs) sit between these extremes, but inverse PINNs typically assume a known governing equation with fixed coefficients, leading to identifiability failures when the true dynamics are unknown or state-dependent. We propose \textbf{SOLIS}, which models unknown dynamics via a \emph{state-conditioned second-order surrogate model} and recasts identification as learning a Quasi-Linear Parameter-Varying (Quasi-LPV) representation, recovering interpretable natural frequency, damping, and gain without presupposing a global equation. SOLIS decouples trajectory reconstruction from parameter estimation and stabilizes training with a cyclic curriculum and \textbf{Local Physics Hints} windowed ridge-regression anchors that mitigate optimization collapse. Experiments on benchmarks show accurate parameter-manifold recovery and coherent physical rollouts from sparse data, including regimes where standard inverse methods fail.
\end{abstract}


\IEEEpeerreviewmaketitle

\section{Introduction}
System Identification (SysID) connects observed data to models that can be analyzed and used for control synthesis \cite{pillonetto2025survey}. A persistent tension is between \emph{interpretability} and \emph{flexibility}. Classical SysID pipelines yield structured and control-oriented models, but they can become restrictive when the dynamics are strongly nonlinear. At the other extreme, modern deep learning approaches offer highly expressive dynamical models \cite{dai2024deep}. In particular, Neural Ordinary Differential Equations (NODEs) learn continuous-time vector fields from data \cite{chen2018neural, ferah2025introducing}, yet the learned representation is typically unstructured, making it difficult to extract control-oriented quantities such as damping ratios, natural frequencies, or stability margins.

Physics-Informed Neural Networks (PINNs) partially address this by embedding differential constraints into the learning objective \cite{raissi2019physics}. For SysID, Inverse PINN (IPINN) training can reconstruct trajectories and estimate parameters from sparse or irregular measurements without explicit numerical integration \cite{raissi2019physics}. In practice, however, training is often ill-conditioned and sensitive to loss balancing and gradient pathologies, which can lead to convergence to poor solutions or outright failure \cite{wang2021gradpath,wang2022ntk,krishnapriyan2021failure}. Beyond optimization, IPINN settings face identifiability issues when coefficients are underdetermined or effectively state-dependent, motivating explicit analyses of identifiability and predictability \cite{kharazmi2021identifiability}. Related modeling families reflect the same trade-off: discrete-time sequence models (RNN/GRU/LSTM) can predict well but hide dynamics in latent states \cite{hochreiter1997lstm,cho2014gru,dong2023nnpsid,pillonetto2025survey}, and neural state-space models improve expressiveness and uncertainty handling but typically do not expose classical parameters in a directly usable form \cite{gedon2021deepSSM,lin2024ssmreview, sertbacs2025stable}. Continuous-time neural modeling, including latent NODE variants, accommodates irregular sampling and has motivated practical identification pipelines for continuous-time state-space models \cite{chen2018neural,rubanova2019latentode,forgione2021ctsyd,beintema2023subnet}, yet mapping a generic learned vector field into control-relevant structure remains nontrivial even when rollouts are accurate. Hybrid “hidden-physics” approaches attempt to learn unknown components while enforcing known structure \cite{raissi2018deephpm}, but they inherit the same optimization fragilities \cite{wang2021gradpath,wang2022ntk,krishnapriyan2021failure} and identifiability limits highlighted in applied studies \cite{stiasny2020pinn_powersys,lai2021pino_de}.

We propose \textbf{SOLIS} (Second-Order Local Identification of Surrogates) to bridge data-driven flexibility with physically interpretable surrogate models of nonlinear systems. Rather than learning an unconstrained black-box vector field, SOLIS identifies a \emph{state-conditioned second-order surrogate} that models local trajectory geometry as a mass--spring--damper system with state-varying coefficients. This yields an interpretable Quasi-Linear Parameter-Varying (Quasi-LPV) representation suitable for gain scheduling and stability analysis, while retaining the physics-regularized benefits of IPINN-style learning.

SOLIS is a structured framework that decouples trajectory approximation from physical-law identification and is designed to mitigate the coupled non-convex failure modes reported in inverse settings \cite{wang2021gradpath,wang2022ntk,krishnapriyan2021failure}. Our contributions are threefold:
\begin{enumerate}
    \item \textit{Quasi-LPV Surrogate Formulation:} We define nonlinear identification as learning a state-dependent affine second-order model, mapping complex nonlinearities to interpretable scalar fields.
    
    \item \textit{Solution-Parameter Decomposition:} We introduce a two-network architecture: a \emph{Solution Network} reconstructs trajectories from sparse measurements, while a \emph{Parameter Network} identifies the state-conditioned physics.
    
    \item \textit{Curriculum Learning:} We stabilize the coupled optimization via a cyclic curriculum and sliding-window ridge regression, which provides analytical parameter anchors during early training.
\end{enumerate}
To show the superiority of SOLIS, we conduct comparative experiments on parameter-manifold recovery and physically consistent rollout performance across benchmark systems.
\clearpage

\section{Preliminaries on PINN}
\label{sec:preliminaries}
PINNs provide a framework for solving forward differential equation problems by embedding physical constraints into the learning process. In the standard PINN setting, we assume the system dynamics are governed by a \emph{known} model structure $\mathbf{\hat{f}}$:
\begin{equation}
    \dot{\mathbf{x}}(t) = \mathbf{\hat{f}}(\mathbf{x}(t), u(t)).
    \label{eq:pinn_model}
\end{equation}
To approximate the solution of \eqref{eq:pinn_model}, PINNs employ a neural network $\mathbf{\hat{x}}(t) = \mathcal{N}(t; \mathbf{W})$, parametrized by weights $\mathbf{W}$, which maps time $t$ to the state estimates.

The network is trained to satisfy both the measurement data and the differential equation. This is achieved by minimizing a composite loss function:
\begin{equation}
    \min_{\mathbf{W}} \left( \mathcal{L}_{data}(\mathbf{W}) + \lambda_p \mathcal{L}_{phys}(\mathbf{W}) \right).
\end{equation}
Here, $\mathcal{L}_{data}$ enforces data fidelity, while $\mathcal{L}_{phys}$ acts as a physics-consistency proxy by taking the mean squared $\ell_2$ discrepancy, evaluated over collocation points $t \in \mathcal{T}_c$, between the predicted time derivative $\dot{\hat{\mathbf{x}}}(t)$ and the model right-hand side $\hat{\mathbf{f}}(\hat{\mathbf{x}}(t),u(t))$. Crucially, the time derivatives are computed via automatic differentiation on the computation graph of $\mathcal{N}$, allowing for mesh-free gradient-based optimization.

% \subsubsection{IPINN}
IPINNs extend the PINN framework to SysID tasks by parametrizing the model structure $\hat{f}$ with $\boldsymbol{\theta}$:
\begin{equation}
    \dot{\mathbf{x}}(t) = \mathbf{\hat{f}}(\mathbf{x}(t), u(t); \boldsymbol{\theta}).
    \label{eq:ipinn_model}
\end{equation}
The goal of IPINNs is to jointly estimate the state trajectory $\mathbf{\hat{x}}(t)$ and the system parameters $\boldsymbol{\theta}$. This is formulated as a simultaneous optimization problem:
\begin{equation}
    \min_{\mathbf{W}, \boldsymbol{\theta}} \left( \mathcal{L}_{data}(\mathbf{W}) + \lambda_p \mathcal{L}_{phys}(\mathbf{W}, \boldsymbol{\theta}) \right).
\end{equation}
Thus, the IPINN discovers the parameters $\boldsymbol{\theta}$ that best explain the observed data while maintaining consistency with \eqref{eq:ipinn_model}.

\begin{figure*}[t]
\centering
\begin{tikzpicture}[
    node distance=1.5cm and 1.2cm,
    >=Latex,
    font=\small\sffamily,
    % Styles
    net/.style={
        draw=netborder,
        fill=netblue,
        rounded corners,
        minimum width=2.9cm,
        minimum height=1.25cm,
        align=center,
        thick
    },
    data/.style={
        draw=none,
        fill=none,
        align=center
    },
    loss/.style={
        draw=lossborder,
        circle,
        inner sep=2pt,
        minimum size=1.15cm,
        align=center,
        thick,
        font=\footnotesize\bfseries
    },
    forward/.style={
        ->,
        thick,
        color=black!80
    },
    gradient/.style={
        ->,
        dashed,
        thick
    }
]

% =========================
% Nodes
% =========================
\node[data] (inputs) {%
\textbf{Inputs}\\
$\mathcal{T}_m,\ \mathcal{T}_c,\ \mathbf{x}_0$\\
$u(\mathcal{T}_m),\ u(\mathcal{T}_c)$%
};

\node[net, right=0.75cm of inputs] (solnet) {%
\footnotesize{Solution Network}\\
$\mathcal{N}_{sol}$%
};

\node[data, right=0.75cm of solnet] (states) {%
\textbf{State Estimates}\\
$\hat{\mathbf{x}}(\mathcal{T}_m)=\{\hat{y}_m,\hat{v}_m\}$\\
$\hat{\mathbf{x}}(\mathcal{T}_c)=\{\hat{y}_c,\hat{v}_c\}$%
};

\node[net, right=0.75cm of states] (paramnet) {%
\footnotesize{Parameter Network}\\
$\mathcal{N}_{param}$%
};

\node[data, right=1.25cm of paramnet] (params) {%
\textbf{Affine Coeffs}\\
$\hat{\boldsymbol{\theta}}(\hat{\mathbf{x}}(\mathcal{T}_c))$\\
$\{\hat{k},\hat{d},\hat{g}\}$%
};

% --- Ridge Calculation block (Phase 2 / Orange) ---
\node[net, above=0.85cm of paramnet, fill=p2color!15, draw=p2color] (ridgecalc)
{\footnotesize{Coefficient Ridge Hint}\\
$\theta^*=(\Phi^\top\Phi+\lambda_r I)^{-1}\Phi^\top Y$};

% =========================
% Losses
% =========================
\node[loss, above=0.9cm of solnet, fill=p1color!20, draw=p1color] (ldata) {$\mathcal{L}_{data}$};
\node[data, left=0.95cm of ldata] (groundtruth) {%
\textbf{Measurement Data}\\
$\mathbf{x}_m(\mathcal{T}_m)$%
};

\node[loss, below=1.9cm of states, fill=bothcolor!20, draw=bothcolor] (lphys) {$\mathcal{L}_{phys}$};

% L_hint: add simple ridge relation (small, no overfill)
\node[loss, right=1.03cm of ridgecalc, fill=p2color!20, draw=p2color] (lhint)
{$\mathcal{L}_{hint}$\\[-1pt]\tiny{$\;\theta\approx\theta^*$}\\\tiny{+ $\mathcal{L}_{TV}$}\\\tiny{+ $\mathcal{L}_{step}$}};

% =========================
% Forward flow
% =========================
\draw[forward] (inputs) -- (solnet);
\draw[forward] (solnet) -- (states);
\draw[forward] (states) -- (paramnet) node[pos=0.4, below, font=\tiny] {$\hat{\mathbf{x}}_c$};
\draw[forward] (paramnet) -- (params);

% --- Data loss connections (Phase 1 / Blue) ---
\draw[forward, color=p1color] (groundtruth.east) -- (ldata.west);
% State Est -> L_data (using measurement subset)
\draw[forward, color=p1color] ($(states.north)+(-0.12,0)$) |- (ldata.east) node[pos=0.7, below, font=\tiny] {$\hat{\mathbf{x}}_m$};

% --- Ridge path (Phase 2 / Orange) ---
\draw[forward, color=p2color] ($(states.north)+(0.12,0)$) |- (ridgecalc.west) node[pos=0.7, below, font=\tiny] {$\hat{\mathbf{x}}_c,u_c$};
\draw[forward, color=p2color] (ridgecalc.east) -- (lhint.west);

% --- Physics loss inputs (Both / Violet) ---
\draw[forward, color=bothcolor] (states.south) -- (lphys.north) node[pos=0.5, right, font=\tiny] {$\hat{\mathbf{x}}_c,u_c$};
\draw[forward, color=bothcolor] (params) |- (lphys.east);

% --- Phase 2 hint/regularizer uses parameters (Orange) ---
\draw[forward, color=p2color] (params.north) -- (lhint.south);

% --- Optional bypass arrow for u (kept subtle) ---
% \draw[forward, rounded corners, color=black!60] (inputs.south) -- ++(0,-0.65) -| ($(states.south west) + (0.45, 0)$);

% =========================
% Gradients / Updates
% =========================
% L_data -> Solution (straight)
\draw[gradient, color=p1color] (ldata.south) -- (solnet.north);

% L_phys -> both networks
\draw[gradient, color=bothcolor] (lphys.140) -| (solnet.south);
\draw[gradient, color=bothcolor] (lphys.40) -| ($(paramnet.south west) + (0.4, 0)$);

% L_hint -> Parameter network (custom routed to avoid overlaps):
% start at bottom of lhint with slight xshift, go down, then left to above paramnet,
% then drop into paramnet from its north.
\coordinate (h0) at ($(lhint.south)-(0.18,0)$);
\coordinate (h1) at ($(h0)+(0,-0.2)$);
\coordinate (h2) at ($(paramnet.north)+(0,0.45)$);
\draw[gradient, color=p2color]
    (h0) -- (h1)  -- (h2) -- (paramnet.north);

% =========================
% Background box
% =========================
\begin{scope}[on background layer]
    \node[fit=(solnet)(paramnet)(states)(lphys), fill=gray!5, rounded corners, draw=gray!20, dashed, inner sep=0.3cm] (box) {};
    \node[above right, font=\scriptsize\bfseries\color{gray}] at (box.south west) {Coupled Physics Loop};
\end{scope}

\end{tikzpicture}
\caption{Overview of the SOLIS architecture. The solution network reconstructs continuous-time state trajectories at measurement times $\mathcal{T}_m$ and collocation times $\mathcal{T}_c$. The data loss uses only $\hat{\mathbf{x}}(\mathcal{T}_m)$, whereas the physics, ridge, and Phase~2 regularization terms are enforced using $\hat{\mathbf{x}}(\mathcal{T}_c)$ and the induced coefficients $\hat{\boldsymbol{\theta}}(\hat{\mathbf{x}}(\mathcal{T}_c))$. Loss terms are color-coded by phase: \textbf{\textcolor{orange!90!black}{data loss}} (Phase~1), \textbf{\textcolor{violet!90!black}{hint/regularization losses}} (Phase~2), and the \textbf{\textcolor{red!80!black}{physics loss}} coupling both networks.}
\label{fig:overview}
\end{figure*}
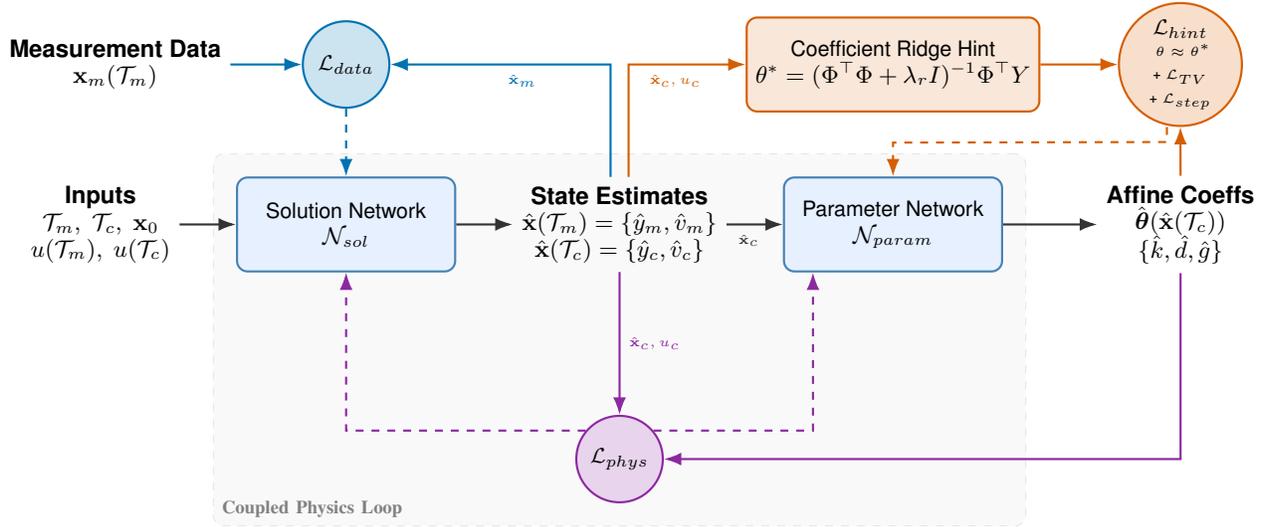

\section{Identification Objective and Hypothesis}
\label{sec:problem}

We consider the problem of identifying a nonlinear dynamical system from sparse and noisy measurements. The underlying system is assumed to evolve based on
\begin{equation}
    \dot{\mathbf{x}}(t) = \mathbf{f}(\mathbf{x}(t), u(t)),
    \label{eq:general_ode}
\end{equation}
where $\mathbf{x}(t) \in \mathbb{R}^n$ denotes the system state, $u(t)$ is a known exogenous input, and the functional form of $\mathbf{f}(\cdot)$ is unknown. Neither the system order nor an explicit parametric structure is assumed a priori.

\subsection{Identification Objective and Setting}

The objective of this work is not to recover the true governing equations of the system, but to infer a data-consistent and physically meaningful surrogate model that explains the observed system trajectories and yields interpretable parameters suitable for control-oriented analysis. 
% To this end, trajectory reconstruction and model identification are treated as coupled learning problems coordinated through physics-informed constraints, enabling the discovery of locally valid dynamical representations.

To achieve this goal, we assume to have measurements from $J (j = 1, \dots, J)$ trajectories. The data is composed of:
\begin{itemize}
    \item A known exogenous input signal $u^{(j)}(t)$, available as a continuous-time function over the horizon $t \in [0, T^{(j)}]$.
    \item Output measurements
    $\mathcal{D}_m^{(j)} = \{(t_i^{(j)}, \mathbf{y}_i^{(j)})\}_{i=1}^{N_d^{(j)}}$,
    where $\mathbf{y}_i^{(j)}$ denotes the observed system states at time $t_i^{(j)}$.
\end{itemize}
The complete measurement dataset is
$\mathcal{D}_m = \bigcup_{j=1}^J \mathcal{D}_m^{(j)}$. Besides, for each trajectory, we assume to have a dense set of collocation points $\mathcal{T}_c^{(j)} = \{t_k^{(j)}\}_{k=1}^{N_c^{(j)}}, \text{with } N_c^{(j)} \gg N_d^{(j)}$.

\subsection{The Second-Order Surrogate Hypothesis}
\label{sec:hypothesis}

Applying physics-informed learning to the problem \eqref{eq:general_ode} presents a fundamental challenge. While standard IPINNs are effective when the exact mathematical structure of $\mathbf{f}$ is known, they face significant difficulties when the model structure is unknown or when the effective parameters vary dynamically across the state space. To bridge this gap, we introduce the \textbf{Second-Order Surrogate Hypothesis}:

\vspace{3pt}
\noindent \emph{The local temporal geometry of smooth system trajectories can be approximated by a second-order Quasi-LPV surrogate dynamics.}
\vspace{3pt}

\noindent Accordingly, we define the surrogate state as $\mathbf{x}(t) = [y(t), v(t)]^\top$, where $y(t)$ is the output and $v(t) = \dot{y}(t)$ is the (potentially latent) velocity. We postulate that the approximating physics model $\mathbf{\hat{f}}$ takes the following affine structure:
\begin{equation}
    \mathbf{\hat{f}}\big(\mathbf{x}, u; \boldsymbol{\theta}(\mathbf{x})\big) = 
    \begin{bmatrix} 
    v \\ 
    -k(\mathbf{x})y - d(\mathbf{x})v + g(\mathbf{x})u 
    \end{bmatrix}.
    \label{eq:surrogate_f}
\end{equation}

In contrast to standard IPINNs where $\boldsymbol{\theta}$ is a constant vector, here the parameter set is defined as a \emph{vector field} conditioned on the state:
$\boldsymbol{\theta}(\mathbf{x}) = [k(\mathbf{x}), d(\mathbf{x}), g(\mathbf{x})]^\top$. This reframes identification as learning a state-conditioned coefficient field $\boldsymbol{\theta}(\mathbf{x})$ that renders the observed trajectories dynamically consistent under the surrogate in \eqref{eq:surrogate_f}. The Quasi-LPV formulation preserves control interpretability at each state $\mathbf{x}$ as the learned coefficients map to canonical second-order quantities, namely natural frequency, damping ratio, and DC gain,:  
\begin{equation}
    \omega_n(\mathbf{x}) = \sqrt{k(\mathbf{x})}, \quad
    \zeta(\mathbf{x}) = \frac{d(\mathbf{x})}{2\sqrt{k(\mathbf{x})}}, \quad
    K(\mathbf{x}) = \frac{g(\mathbf{x})}{k(\mathbf{x})},
    \label{eq:canonical_params}
\end{equation}
 enabling local, state-dependent analysis.
 
\section{Quasi-LPV Surrogate Modeling via Physics-Informed Learning: SOLIS}
\label{sec:methodology}
The proposed formulation creates an intrinsically coupled learning problem. The coefficient fields $k(\mathbf{x}), d(\mathbf{x}), g(\mathbf{x})$ define the differential constraints, but they must be evaluated at the system state $\mathbf{x}(t)$. However, the true state trajectory is unknown (due to sparsity and noise) and must be reconstructed simultaneously. Accordingly, the objective is to jointly learn:
\begin{enumerate}
    \item A continuous-time reconstruction of the trajectories $\mathbf{\hat{x}}^{(j)}(t)$ for all $j=1\dots J$.
    \item A shared parameter network mapping $\mathbf{x} \mapsto [k, d, g]^\top$ that renders the reconstructed trajectories physically consistent according to \eqref{eq:surrogate_f}.
\end{enumerate}

To resolve the intrinsic coupling, we adopt a two-network architecture trained under a two-phase curriculum. As shown in Fig.~\ref{fig:overview}, the framework comprises two interacting neural components: (i) a \emph{solution network} $\mathcal{N}_{sol}$ that reconstructs continuous-time state trajectories, and (ii) a \emph{parameter network} $\mathcal{N}_{param}$ that identifies a state-conditioned affine surrogate of the dynamics. 

These components are coupled through a physics-based consistency loss enforcing the surrogate dynamics along the reconstructed trajectory. The learning process alternates between trajectory reconstruction and state-conditioned parameter identification, stabilizing optimization while enforcing consistency with the surrogate dynamics.

\subsection{Solution Network ($\mathcal{N}_{sol}$)}
This  network approximates the continuous-time state trajectory
$\hat{\mathbf{x}}(t) = [\hat{y}(t), \hat{v}(t)]^\top$.
Unlike standard PINNs that fit a single time-series, $\mathcal{N}_{sol}$ is trained to represent a family of trajectories corresponding to different initial conditions and input profiles. To achieve this, the network is conditioned on a compact representation of the input through a recurrent context embedding and feature-wise modulation.

\subsubsection{Context Embedding}
For each trajectory, the control input
$\mathbf{u}_{c} = \{u(t)\}_{t \in \mathcal{T}^{(m)}_{c}}$
is encoded into a fixed-dimensional context vector $\mathbf{c}$ using a Gated Recurrent Unit (GRU) \cite{cho2014gru}:
\begin{equation}
    \mathbf{c} = \text{GRU}(\mathbf{u}_{c}; \mathbf{W}_{context}).
\end{equation}
Rather than performing sequential state integration, the GRU is used strictly to embed the discrete exogenous input profile into a static context, preserving the continuous-time nature of the solution network.

\subsubsection{FiLM Conditioning}
The context embedding $\mathbf{c}$ modulates the internal feature maps of the solution network via Feature-wise Linear Modulation (FiLM) \cite{perez2018film}. For the $l$-th hidden layer with pre-activation $\mathbf{z}_l$, the modulation is
\begin{equation}
    \mathbf{z}'_l = \boldsymbol{\gamma}_l(\mathbf{c}) \odot \mathbf{z}_l + \boldsymbol{\beta}_l(\mathbf{c}),
\end{equation}
where $\boldsymbol{\gamma}_l$ and $\boldsymbol{\beta}_l$ are learned affine functions of $\mathbf{c}$. This conditioning enables a single shared network to represent multiple trajectory geometries while preserving a consistent temporal coordinate system.

\subsubsection{Temporal Fourier Encoding (Optional)}
To capture high-frequency oscillatory dynamics for complex systems, we optionally use Random Fourier Features (RFF) \cite{tancik2020fourier}. The scalar time is mapped via $ \mathbf{e}(t) = \left[ \cos(2\pi \mathbf{B}t), \; \sin(2\pi \mathbf{B}t) \right]^\top,$
where $\mathbf{B}$ is a fixed vector of frequencies sampled from ${N}(0, \sigma^2)$.

\subsection{Parameter Network ($\mathcal{N}_{param}$)}
The parameter network maps the instantaneous state and input to the affine surrogate coefficients:
\begin{equation}
    [\hat{k}, \hat{d}, \hat{g}]^\top
    = \mathcal{N}_{param}(\hat{y}, \hat{v}, u; \mathbf{W}_{param}).
\end{equation}
Since $\mathcal{N}_{\mathrm{param}}$ is conditioned on $\mathbf{\hat{x}}$, physical consistency is enforced across trajectories, ensuring that the learned parameter field represents a global, state-conditioned model.

\subsubsection{Mixture-of-Experts (Optional)}
For systems exhibiting highly nonlinear or regime-dependent dynamics, the parameter network may optionally employ a Mixture-of-Experts (MoE) head to improve local expressivity. In this case, the surrogate coefficients are computed as
\begin{equation}
    [\hat{k}, \hat{d}, \hat{g}]^\top
    = \sum_{j=1}^{M} \alpha_j(\mathbf{x}, u)\,\mathcal{E}_j(\mathbf{x}, u),
\end{equation}
where the mixing weights $\boldsymbol{\alpha}(\mathbf{x}, u)$ are produced by a Softmax gating network. For simpler systems, a standard feed-forward parameter network is used.

\subsubsection{State Feature Augmentation (Optional)}
To accelerate the discovery of complex parameter manifolds, we optionally augment the input space of $\mathcal{N}_{param}$ with fixed nonlinear transformations. The raw state vector is expanded to include polynomial terms (e.g., $y^2, y^3$), absolute values, and interaction terms (e.g., $y \cdot v$). This simple feature engineering provides a strong inductive bias for complex physical systems.

\subsection{Curriculum Training Strategy}
\label{sec:curriculum}

Naïve joint optimization of the solution and parameter networks creates a highly ill-conditioned landscape, often collapsing to trivial solutions (e.g., $\hat{y}\to 0, \hat{k}\to 0$) due to gradient conflict. To stabilize training, we employ a \emph{cyclic curriculum} that alternates between two distinct phases (see Algorithm~\ref{alg:SOLIS}\footnote[1]{Python implementation: \url{https://github.com/Assaciry/solis}}). This strategy decouples the learning of trajectory geometry from the discovery of the parameter manifold, preventing early-stage noise from corrupting the physics identification. 
\begin{itemize}
    \item \textbf{Phase 1}-Trajectory Reconstruction:
We freeze $\mathcal{N}_{param}$ and optimize only $\mathcal{N}_{sol}$. The objective is to recover a smooth continuous-time trajectory that fits the sparse measurements. 
\begin{equation}
\mathcal{L}_{\mathrm{P1}}
= \lambda_d \mathcal{L}_{data}
+ \lambda_{ic} \mathcal{L}_{ic}
+ \lambda_p \mathcal{L}_{phys}.
\end{equation}
The physics residual acts as a \emph{geometric regularizer}, guiding the interpolation to respect the differential constraints induced by the current parameter estimates.
\item \textbf{Phase 2}-Parameter Identification: 
We freeze $\mathcal{N}_{sol}$ and optimize $\mathcal{N}_{param}$. 
\begin{equation}
\begin{aligned}
\mathcal{L}_{\mathrm{P2}}
&= \lambda_p \mathcal{L}_{phys}
+ \lambda_{roll} \mathcal{L}_{rollout}
+ \lambda_h \mathcal{L}_{hint}
\\
&+\lambda_{reg} (\mathcal{L}_{TV} + \mathcal{L}_{div}).
\end{aligned}
\end{equation}
Here, to prevent optimization collapse in early iterations, we introduce the  \textbf{Local Physics Hint} loss $\mathcal{L}_{hint}$. It anchors the network to analytical parameter estimates $\boldsymbol{\theta}^{*}_{ridge}$ derived from sliding-window ridge regression, providing a convex "warm-start" for the non-convex physics loss.
\end{itemize}
Phases~1 and~2 are executed in an alternating loop to promote mutual convergence. 

Each phase employs a phase-specific composite loss aligned with its learning objective. The remainder of this section details the individual loss components.

\subsubsection{Physics Consistency (Phase 1 \& 2)}
The core coupling term enforces the surrogate dynamics at all collocation points:
\begin{equation}
    \mathcal{L}_{phys} = \frac{1}{|\mathcal{T}_{c}|} \sum_{t \in \mathcal{T}_{c}} \bigl\| \dot{\hat{v}}(t) + \hat{k}\,\hat{y}(t) + \hat{d}\,\hat{v}(t) - \hat{g}\,u(t) \bigr\|_2^2,
\end{equation}

\subsubsection{Trajectory Fidelity (Phase 1)}
To anchor $\mathcal{N}_{sol}$, we impose consistency with measurements and initial conditions:
\begin{equation}
\begin{aligned}
    \mathcal{L}_{data} &= \frac{1}{|\mathcal{D}_{m}|} \sum_{(t, \mathbf{x}_{m}) \in \mathcal{D}_{m}} \bigl\| \hat{\mathbf{x}}_m(t) - \mathbf{x}_{m}(t) \bigr\|_2^2, \\
    \mathcal{L}_{ic} &= \bigl\| \hat{\mathbf{x}}_0 - \mathbf{x}_{0} \bigr\|_2^2.
\end{aligned}
\end{equation}
\begin{algorithm}[!t]
\caption{SOLIS Training Procedure}
\label{alg:SOLIS}
\begin{algorithmic}[1]
\State \textbf{Input:} Measurements $\mathcal{D}_m$, Collocation points $\mathcal{D}_c$, Phase lengths $K_{1,2}$, Hint decay $\gamma$, Ridge $\lambda_r$
\State \textbf{Initialize:} Networks $\mathcal{N}_{sol}, \mathcal{N}_{param}$  

\For{$epoch = 1$ to $E$}
    \State Sample minibatch $\mathcal{B}$ from $\mathcal{D}_m$ and $\mathcal{D}_c$
    \State $\hat{\mathbf{x}}, \dot{\hat{\mathbf{x}}} \leftarrow \mathcal{N}_{sol}(\mathcal{B})$ \Comment{Forward \& Autograd}
    \State $\hat{\boldsymbol{\theta}} \leftarrow \mathcal{N}_{param}(\hat{\mathbf{x}}, \mathbf{u})$ \Comment{Parameter prediction}
    
    \State Compute $\mathcal{L}_{phys}$
    
    \State $k \leftarrow epoch \bmod (K_1 + K_2)$
    
    \If{$k < K_1$} \Comment{Phase 1: Trajectory Learning}
        \State Freeze $\mathcal{N}_{param}$, Unfreeze $\mathcal{N}_{sol}$
        \State Compute $\mathcal{L}_{data}$ and $\mathcal{L}_{ic}$
        \State $\mathcal{L}_{total} \leftarrow \lambda_d\mathcal{L}_{data} + \lambda_{ic}\mathcal{L}_{ic} + \lambda_p\mathcal{L}_{phys}$
        \State \textbf{Step} $\mathcal{N}_{sol}$ optimizer
        
    \Else \Comment{Phase 2: Parameter Identification}
        \State Freeze $\mathcal{N}_{sol}$, Unfreeze $\mathcal{N}_{param}$
        \State Decay hint weight: $\lambda_h \leftarrow \lambda_h \cdot \gamma$
        
        \State \textbf{Physics Hint:} 
        \State \quad Construct matrices $\mathbf{\Phi}, \mathbf{Y}$ locally on batch
        \State \quad Compute $\boldsymbol{\theta}^*_{ridge}$ via Eq.~\eqref{eq:ridge_sol}
        \State \quad Compute $\mathcal{L}_{hint}$ and weights $w_t$
        
        \State Compute $\mathcal{L}_{roll}$ and $\mathcal{L}_{TV}$
        \State $\mathcal{L}_{total} \leftarrow \lambda_p\mathcal{L}_{phys} + \lambda_h\mathcal{L}_{hint} + \lambda_{reg}(\mathcal{L}_{roll} + \mathcal{L}_{TV})$
        \State \textbf{Step} $\mathcal{N}_{param}$ optimizer
    \EndIf
\EndFor
\State \textbf{Return} Trained networks $\mathcal{N}_{sol}$ and  $\mathcal{N}_{param}$
\end{algorithmic}
\end{algorithm}

\subsubsection{Local Ridge Hints (Phase 2)}
To prevent parameter collapse early in training, we compute a local analytical estimate $\boldsymbol{\theta}^*_{ridge}(t)$ using a \emph{random rolling window} scheme. At each optimization step, we sample an odd window length $w \sim \mathcal{U}\{5,|\mathcal{T}_c|\}$ and form, for every $t \in \mathcal{T}_c$, a local window $W_t^{(w)}$ of size $w$ extracted by a stride-1 rolling. Within each window, the dynamics are approximated as 
$\mathbf{Y} \approx \mathbf{\Phi}\boldsymbol{\theta}$:
\begin{equation}
    \mathbf{Y} = [\dot{\hat{v}}(\tau)]_{\tau \in W_t^{(w)}}, \   
    \mathbf{\Phi} = [-\hat{y}(\tau), -\hat{v}(\tau), u(\tau)]_{\tau \in W_t^{(w)}}.
\end{equation}
The corresponding ridge estimator is computed in closed form,
\begin{equation}
    \boldsymbol{\theta}^*_{ridge}(t) = (\mathbf{\Phi}^\top \mathbf{\Phi} + \lambda_r \, \mathbf{I})^{-1}\mathbf{\Phi}^\top\mathbf{Y},
    \label{eq:ridge_sol}
\end{equation}
where $\lambda_r$ may be constant or scaled with the local feature energy (i.e., adaptive ridge) to avoid over-regularization in low-excitation windows.

The hint loss penalizes deviation from this analytical proxy, weighted by a local reliability score:
\begin{equation}
    \mathcal{L}_{hint} = \frac{1}{|\mathcal{T}_{c}|} \sum_{t \in \mathcal{T}_{c}} w_t \bigl\| \boldsymbol{\hat{\theta}}(t) - \boldsymbol{\theta}^*_{ridge}(t) \bigr\|_2^2.
\end{equation}
The weight $w_t \in [0,1]$ is derived from the eigen-structure of $\mathbf{\Phi}^\top\mathbf{\Phi}$, which suppresses hints in ill-conditioned or weakly excited regimes. Randomizing the window length across iterations prevents the hint from locking onto a single temporal scale and encourages $\boldsymbol{\hat{\theta}}(t)$ to remain consistent under multiple local linearizations, improving stability in Phase~2.

\subsubsection{Parameter Regularization (Phase 2)}
To ensure temporal consistency, we employ Total Variation (TV) regularization on the parameter field and a short-horizon ($H$-step) rollout loss:
\begin{equation}   
\begin{aligned}
    \mathcal{L}_{TV} &= \frac{1}{|\mathcal{T}_{c}|} \sum_{t} \bigl\| \nabla_x \boldsymbol{\hat{\theta}}(\hat{\mathbf{x}}_t) \bigr\|_1, \\
    \mathcal{L}_{roll} &= \frac{1}{H} \sum_{h=1}^{H} \bigl\| \text{RK4}(\hat{\mathbf{x}}_t, \hat{\boldsymbol{\theta}}_t, \dots) - \hat{\mathbf{x}}_{t+h} \bigr\|_2^2.
\end{aligned}
\end{equation}

\section{Performance Analysis}

In this section, we evaluate the performance of SOLIS on 
\begin{itemize}
    \item Duffing Oscillator (Simulated): A nonlinear benchmark exhibiting cubic stiffness effects, with dynamics governed by a state-dependent stiffness $k(x) = \alpha + \beta x^2$.
    \item Van der Pol Oscillator (Simulated): Assesses the identification of self-sustained oscillations induced by nonlinear, state-dependent damping $d(x) = \mu(x^2 - 1)$.
    \item Two-Tank System (Real dataset): A laboratory-scale fluid process characterized by coupled tank dynamics and nonlinear level–flow relationships under measurement noise.
\end{itemize}
against the following methods: 
\begin{itemize}
    \item \textbf{IPINN:} A baseline inverse PINN assuming constant global parameters ($k, d, g$).
    \item \textbf{IPINN-M:} IPINN with a multi-trajecory setting.
    \item \textbf{TF:} A second-order transfer function estimated with MATLAB's built-in \texttt{tfest} function.
\end{itemize}

Trajectory and rollout performance are quantified via \emph{Accuracy}, defined as $(1 - \text{NRMSE}) \times 100\%$, with NRMSE normalized by the true signal's peak-to-peak range.

\subsection{Solution Reconstruction on Observed Data}
\label{sec:solution_results}

\begin{figure}[t]
    \centering
    \includegraphics[width=1\columnwidth,keepaspectratio]{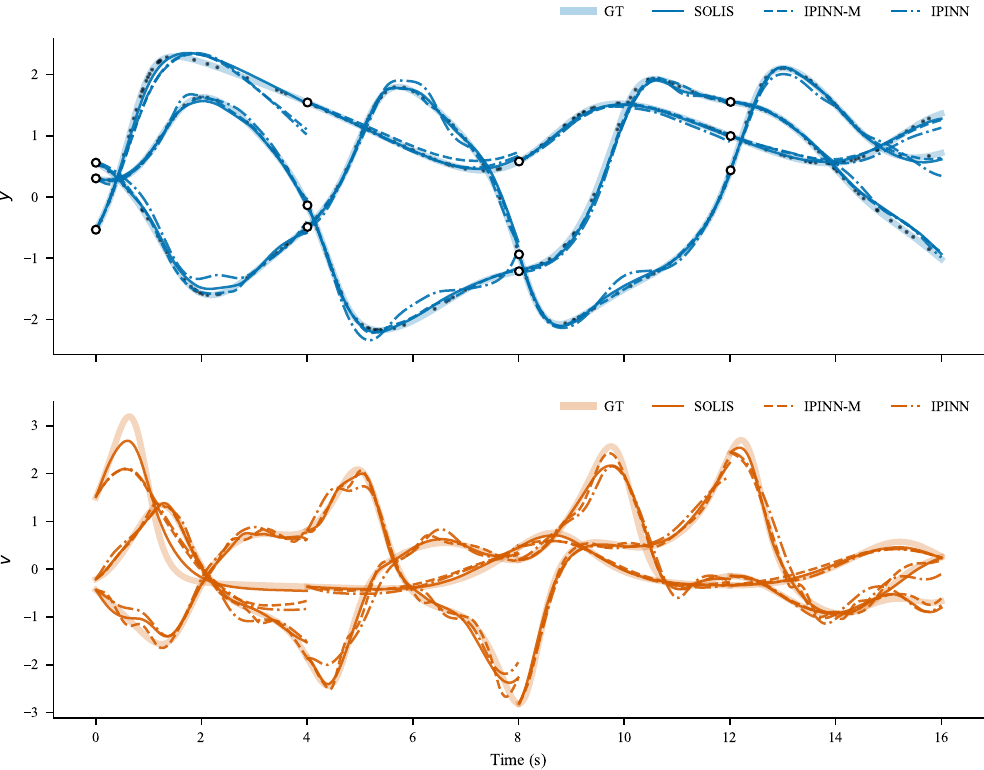}

    \caption{Van der Pol in-sample reconstruction. Top: $y(t)$ (markers) with Ground Truth (GT) and reconstructed trajectories. Bottom: corresponding $v(t)$.}

    \label{fig:solution_results}
\end{figure}

We first evaluate \emph{in-sample} solution reconstruction on the observed trajectories to verify that $\hat{\mathbf{x}}(t)$ accurately fits the measurements while remaining physics consistent.

Fig.~\ref{fig:solution_results} shows a representative Van der Pol training subset. SOLIS closely tracks the ground-truth trajectories, whereas IPINN and IPINN-M exhibit drift and local mismatches, particularly near peaks and high-curvature regions. Table~\ref{tab:solution_reconstruction} confirms this behavior across all training trajectories, with SOLIS achieving the highest reconstruction accuracy.

\subsection{Identification and Generalization}
\label{sec:identification_results}

Here, we evaluate the generative capabilities of the learned surrogate $\mathcal{N}_{param}$, i.e., whether the identified Quasi-LPV dynamics recover the underlying vector field and produce stable open-loop predictions beyond the training windows.

\paragraph{Phase Portrait Comparison}
For Duffing and Van der Pol, we compare the surrogate-induced phase portrait with the ground-truth field. As illustrated in Fig.~\ref{fig:phase_portraits}, SOLIS reproduces the qualitative flow structure and matches the ground-truth directions over most of the explored state space; the cosine-similarity map is highest along the regions covered by the training trajectories. Table~\ref{tab:phase_portrait_similarity} confirms this quantitatively: SOLIS attains the highest average similarity on both systems, indicating more faithful recovery of the underlying vector field.

\paragraph{Predictive Rollout}
To assess generalization, we perform open-loop forward integration on unseen test trajectories. Starting from an initial condition $\mathbf{x}_0$, we integrate the learned Quasi-LPV dynamics using the identified parameter network:

\begin{figure}[t]
\centering

\includegraphics[width=0.98\columnwidth]{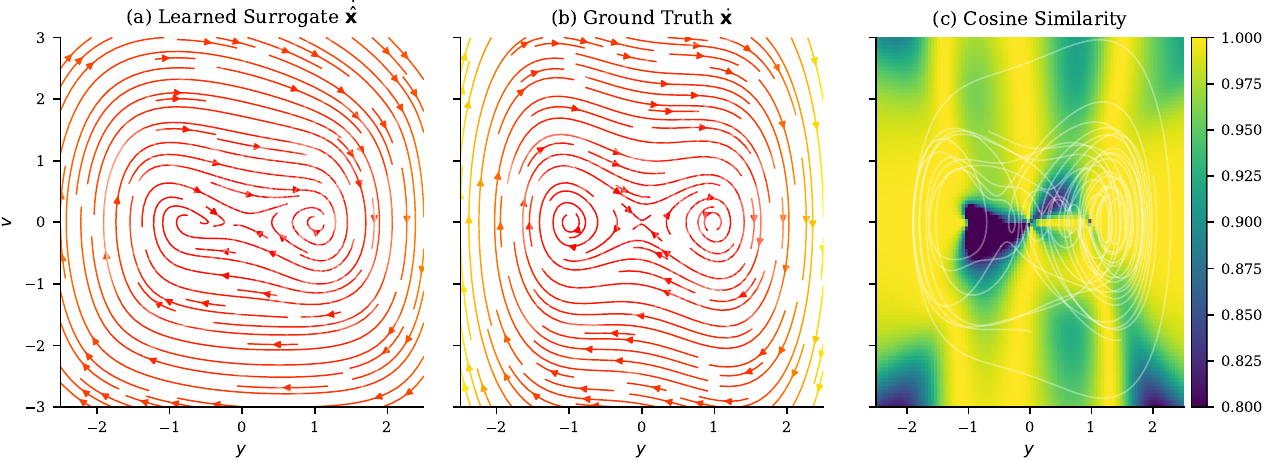}\\[0.8em]

\caption{Duffing Oscillator Phase portrait comparison  (a) Surrogate phase portrait calculated via \eqref{eq:surrogate_state_space}, (b) ground-truth phase portrait, (c) cosine similarity between surrogate and ground-truth vector fields over the state space, with training trajectories shown in white.}
\label{fig:phase_portraits}
\end{figure}
\begin{figure}[t]
\centering

\includegraphics[width=1\columnwidth,keepaspectratio]{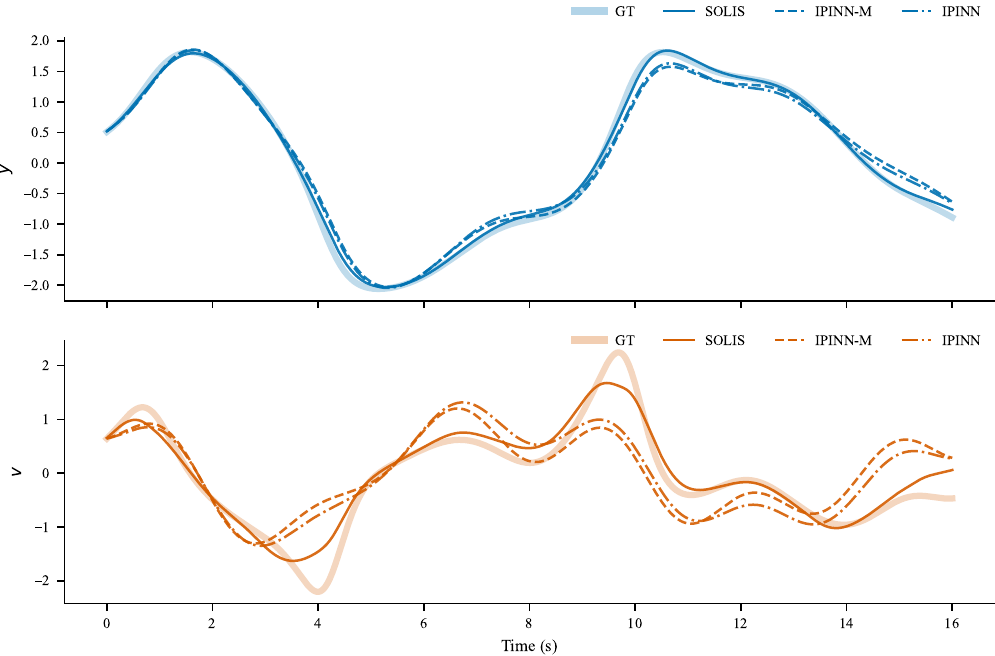}
\caption{Van der Pol oscillator open-loop test rollout. Trajectories obtained by forward integrating the learned surrogate dynamics \eqref{eq:surrogate_state_space} from the same initial condition are compared against the ground truth for both state channels.}

\label{fig:rollout}
\end{figure}

\begin{table}[t]
\caption{Solution Learning Accuracy ($\uparrow$)  on Training Trajectories}
\label{tab:solution_reconstruction}
\centering
\setlength{\tabcolsep}{10pt} % Increased to 10pt so the table isn't too tiny
\renewcommand{\arraystretch}{1.2} 

\begin{tabular}{|l|c|c|c|}
\hline
\textbf{System} & \textbf{IPINN} & \textbf{IPINN-M} & \textbf{SOLIS} \\
\hline
Van der Pol & 95.94\% & 98.07\% & 98.89\% \\
\hline
Duffing     & 95.43\% & 97.56\% & 99.01\% \\
\hline
Two-Tank    & 88.64\% & 91.25\% & 98.62\% \\
\hline
\end{tabular}
\end{table}
\begin{table}[t]
\caption{Phase portrait similarity ($\uparrow$)  measured via average cosine similarity to the ground-truth field.}
\label{tab:phase_portrait_similarity}
\centering
\setlength{\tabcolsep}{10pt} % You can increase this (e.g., 10pt) to add a little more padding if it looks too tight
\renewcommand{\arraystretch}{1.2} 

\begin{tabular}{|l|c|c|c|} % Removed tabular* and the fill command
\hline
\textbf{System} & \textbf{IPINN} & \textbf{IPINN-M} & \textbf{SOLIS} \\
\hline
Van der Pol & 0.70 & 0.67 & 0.86 \\
\hline
Duffing     & 0.81 & 0.82 & 0.96 \\
\hline
\end{tabular}
\end{table}

\begin{equation}
\dot{\hat{\mathbf{x}}} =
\begin{bmatrix}
\hat{v} \\
-\mathcal{N}_{param}(\hat{\mathbf{x}}, u)\cdot[\hat{y}, \hat{v}, -u]^\top
\end{bmatrix}.
\label{eq:surrogate_state_space}
\end{equation}
Fig.~\ref{fig:rollout} shows that SOLIS maintains coherent test-time rollouts that track the reference evolution, whereas the baselines accumulate drift over time. The aggregate results in Table~\ref{tab:identification_rollout} confirm this trend across all benchmarks, with SOLIS achieving the highest accuracy, indicating that its surrogate captures dynamics that generalize beyond in-sample reconstruction.

\begin{table}[t]
\caption{Identification rollout accuracy ($\uparrow$)  on test trajectories.}
\label{tab:identification_rollout}
\centering
\setlength{\tabcolsep}{8pt} 
\renewcommand{\arraystretch}{1.2} 

\begin{tabular}{|l|c|c|c|c|}
\hline
\textbf{System} & \textbf{IPINN} & \textbf{IPINN-M} & \textbf{TF} & \textbf{SOLIS} \\
\hline
Van der Pol & 83.72\% & 81.84\% & 70.22\% & 90.54\% \\
\hline
Duffing     & 76.60\% & 75.27\% & 74.60\% & 83.07\% \\
\hline
Two-Tank    & 82.74\% & 82.9\%  & 77.74\% & 84.29\% \\
\hline
\end{tabular}
\end{table}

% =========================================================================
% 6. CONCLUSION
% =========================================================================
\section{Conclusion and Future Work}
\label{sec:conclusion}

This work proposed SOLIS, a PINN framework for identifying interpretable second-order surrogate models of nonlinear systems. By decomposing the learning problem into trajectory geometry and parameter manifold identification, we addressed the identifiability and stability challenges common in IPINNs.

Our key technical contribution lies in the structured curriculum strategy, specifically the introduction of \emph{Local Physics Hints}. We demonstrated that using a sliding-window ridge regression proxy to warm-start the parameter network acts as an effective convex regularization, preventing the optimization from collapsing into trivial solutions. Furthermore, by framing the interaction between the solution and parameter networks as a self-regulated feedback loop, we established a robust mechanism for discovering physical laws that are consistent with observed data. Experimental results confirm that SOLIS can recover stable, state-dependent parameter profiles even in regimes where global parameter identification fails. 

While our current evaluation validates the framework on fundamental low-dimensional benchmarks, future work will scale this methodology to higher-dimensional and more complex nonlinear systems. Additionally, we plan to extend this framework to multi-input multi-output systems and explore the application of the learned surrogates in predictive control settings.

\section*{Acknowledgment}
The authors acknowledge using ChatGPT and Gemini to refine the grammar and enhance the expression of English.

\bibliographystyle{IEEEtran}
\bibliography{IEEEabrv,cites}

\begin{thebibliography}{10}
\providecommand{\url}[1]{#1}
\csname url@samestyle\endcsname
\providecommand{\newblock}{\relax}
\providecommand{\bibinfo}[2]{#2}
\providecommand{\BIBentrySTDinterwordspacing}{\spaceskip=0pt\relax}
\providecommand{\BIBentryALTinterwordstretchfactor}{4}
\providecommand{\BIBentryALTinterwordspacing}{\spaceskip=\fontdimen2\font plus
\BIBentryALTinterwordstretchfactor\fontdimen3\font minus \fontdimen4\font\relax}
\providecommand{\BIBforeignlanguage}[2]{{%
\expandafter\ifx\csname l@#1\endcsname\relax
\typeout{** WARNING: IEEEtran.bst: No hyphenation pattern has been}%
\typeout{** loaded for the language `#1'. Using the pattern for}%
\typeout{** the default language instead.}%
\else
\language=\csname l@#1\endcsname
\fi
#2}}
\providecommand{\BIBdecl}{\relax}
\BIBdecl

\bibitem{pillonetto2025survey}
G.~Pillonetto, A.~Aravkin, D.~Gedon, L.~Ljung, A.~H. Ribeiro, and T.~B. Sch{\"o}n, ``Deep networks for system identification: A survey,'' \emph{Automatica}, vol. 171, p. 111907, 2025.

\bibitem{dai2024deep}
T.~Dai, K.~Aljanaideh, R.~Chen, R.~Singh, A.~Stothert, and L.~Ljung, ``Deep learning of dynamic systems using system identification toolbox™,'' \emph{IFAC-PapersOnLine}, vol.~58, no.~15, pp. 580--585, 2024.

\bibitem{chen2018neural}
R.~T.~Q. Chen, Y.~Rubanova, J.~Bettencourt, and D.~K. Duvenaud, ``Neural ordinary differential equations,'' in \emph{Advances in Neural Information Processing Systems}, vol.~31, 2018.

\bibitem{ferah2025introducing}
M.~A. Ferah and T.~Kumbasar, ``Introducing interval neural networks for uncertainty-aware system identification,'' in \emph{Int. Congress on Human-Computer Interaction, Optimization and Robotic Applications}, 2025.

\bibitem{raissi2019physics}
M.~Raissi, P.~Perdikaris, and G.~E. Karniadakis, ``Physics-informed neural networks: A deep learning framework for solving forward and inverse problems involving nonlinear partial differential equations,'' \emph{Journal of Computational Physics}, vol. 378, pp. 686--707, 2019.

\bibitem{wang2021gradpath}
S.~Wang, Y.~Teng, and P.~Perdikaris, ``Understanding and mitigating gradient flow pathologies in physics-informed neural networks,'' \emph{SIAM Journal on Scientific Computing}, vol.~43, no.~5, pp. A3055--A3081, 2021.

\bibitem{wang2022ntk}
S.~Wang, X.~Yu, and P.~Perdikaris, ``When and why pinns fail to train: A neural tangent kernel perspective,'' \emph{Journal of Computational Physics}, vol. 449, p. 110768, 2022.

\bibitem{krishnapriyan2021failure}
A.~S. Krishnapriyan, A.~Gholami, S.~Zhe, R.~M. Kirby, and M.~W. Mahoney, ``Characterizing possible failure modes in physics-informed neural networks,'' in \emph{Advances in Neural Information Processing Systems}, vol.~34, 2021.

\bibitem{kharazmi2021identifiability}
E.~Kharazmi, M.~Cai, X.~Zheng, Z.~Zhang, G.~Lin, and G.~E. Karniadakis, ``Identifiability and predictability of integer- and fractional-order epidemiological models using physics-informed neural networks,'' \emph{Nature Computational Science}, vol.~1, pp. 744--753, 2021.

\bibitem{hochreiter1997lstm}
S.~Hochreiter and J.~Schmidhuber, ``Long short-term memory,'' \emph{Neural Computation}, vol.~9, no.~8, pp. 1735--1780, 1997.

\bibitem{cho2014gru}
K.~Cho, B.~van Merri{\"e}nboer, C.~G{\"u}l{\c{c}}ehre, D.~Bahdanau, F.~Bougares, H.~Schwenk, and Y.~Bengio, ``Learning phrase representations using rnn encoder--decoder for statistical machine translation,'' in \emph{Conference on Empirical Methods in Natural Language Processing}, 2014.

\bibitem{dong2023nnpsid}
A.~Dong, A.~Starr, and Y.~Zhao, ``Neural network-based parametric system identification: a review,'' \emph{International Journal of Systems Science}, vol.~54, no.~13, pp. 2676--2688, 2023.

\bibitem{gedon2021deepSSM}
D.~Gedon, N.~Wahlstr{\"o}m, T.~B. Sch{\"o}n, and L.~Ljung, ``Deep state space models for nonlinear system identification,'' \emph{IFAC-PapersOnLine}, vol.~54, no.~7, pp. 481--486, 2021.

\bibitem{lin2024ssmreview}
J.~Lin and G.~Michailidis, ``Deep learning-based approaches for state space models: A selective review,'' arXiv preprint, 2024.

\bibitem{sertbacs2025stable}
A.~E. Sertba{\c{s}} and T.~Kumbasar, ``Stable-by-design neural network-based lpv state-space models for system identification,'' in \emph{Conference of Image Processing, Wavelet and Applications on Real World Problems}, 2025.

\bibitem{rubanova2019latentode}
Y.~Rubanova, T.~Q. Chen, and D.~K. Duvenaud, ``Latent ordinary differential equations for irregularly-sampled time series,'' in \emph{Advances in Neural Information Processing Systems}, vol.~32, 2019, pp. 5321--5331.

\bibitem{forgione2021ctsyd}
M.~Forgione and D.~Piga, ``Continuous-time system identification with neural networks: Model structures and fitting criteria,'' \emph{European Journal of Control}, vol.~59, pp. 69--81, 2021.

\bibitem{beintema2023subnet}
G.~I. Beintema, M.~Schoukens, and R.~T{\'o}th, ``Continuous-time identification of dynamic state-space models by deep subspace encoding,'' in \emph{International Conference on Learning Representations}, 2023.

\bibitem{raissi2018deephpm}
M.~Raissi, ``Deep hidden physics models: Deep learning of nonlinear partial differential equations,'' arXiv preprint, 2018.

\bibitem{stiasny2020pinn_powersys}
J.~Stiasny, G.~S. Misyris, and S.~Chatzivasileiadis, ``Physics-informed neural networks for non-linear system identification for power system dynamics,'' arXiv preprint, 2020.

\bibitem{lai2021pino_de}
Z.~Lai, C.~Mylonas, S.~Nagarajaiah, and E.~Chatzi, ``Structural identification with physics-informed neural ordinary differential equations,'' \emph{Journal of Sound and Vibration}, vol. 508, p. 116196, 2021.

\bibitem{perez2018film}
E.~Perez, F.~Strub, H.~de~Vries, V.~Dumoulin, and A.~Courville, ``Film: Visual reasoning with a general conditioning layer,'' in \emph{Conference on Artificial Intelligence}, vol.~32, no.~1, 2018, pp. 3942--3950.

\bibitem{tancik2020fourier}
M.~Tancik, P.~P. Srinivasan, B.~Mildenhall, S.~Fridovich-Keil, N.~Raghavan, U.~Singhal, R.~Ramamoorthi, J.~T. Barron, and R.~Ng, ``Fourier features let networks learn high frequency functions in low dimensional domains,'' in \emph{Advances in Neural Information Processing Systems}, vol.~33, 2020, pp. 7537--7547.

\end{thebibliography}

\end{document}